\documentclass[twocolumn, switch]{article} % Method A for two-column formatting

\usepackage{preprint}
\usepackage{background}
\backgroundsetup{contents={}}

\usepackage{multirow}
\usepackage{pifont}
\usepackage{graphicx}
\usepackage{float}
\usepackage{afterpage}
\usepackage{caption}
\usepackage{hyperref}
\usepackage{url}

\usepackage{graphicx}
\usepackage{subcaption}

\usepackage{amsmath, amsthm, amssymb, amsfonts}

\usepackage[numbers,square]{natbib}
\usepackage[utf8]{inputenc}	% allow utf-8 input
\usepackage[T1]{fontenc}	% use 8-bit T1 fonts
\usepackage{xcolor}		% colors for hyperlinks
\usepackage{newfloat}
\DeclareFloatingEnvironment[name={Supplementary Figure}]{suppfigure}
\usepackage{sidecap}
\sidecaptionvpos{figure}{c}

\usepackage{titlesec}
\titlespacing\section{0pt}{12pt plus 3pt minus 3pt}{1pt plus 1pt minus 1pt}
\titlespacing\subsection{0pt}{10pt plus 3pt minus 3pt}{1pt plus 1pt minus 1pt}
\titlespacing\subsubsection{0pt}{8pt plus 3pt minus 3pt}{1pt plus 1pt minus 1pt}

\usepackage{tikz,xcolor,hyperref}

\definecolor{lime}{HTML}{A6CE39}
\DeclareRobustCommand{\orcidicon}{
	\begin{tikzpicture}
	\draw[lime, fill=lime] (0,0)
	circle [radius=0.16]
	node[white] {{\fontfamily{qag}\selectfont \tiny ID}};
	\draw[white, fill=white] (-0.0625,0.095)
	circle [radius=0.007];
	\end{tikzpicture}
	\hspace{-2mm}
}
\foreach \x in {A, ..., Z}{\expandafter\xdef\csname orcid\x\endcsname{\noexpand\href{https://orcid.org/\csname orcidauthor\x\endcsname}
			{\noexpand\orcidicon}}
}
\title{STAGNet: A Spatio-Temporal Graph and LSTM Framework for Accident Anticipation}

\usepackage{geometry}
\usepackage{eso-pic}
\usepackage{xcolor}

\AddToShipoutPictureBG*{%
  \AtPageLowerLeft{%
    \put(\LenToUnit{0.7in},\LenToUnit{0.5in}){% % Left-aligned line
      \rule{2in}{0.5pt}}%
    \put(\LenToUnit{0.82in},\LenToUnit{0.33in}){% % Left-aligned text
      \raggedright\textcolor{black}{\normalfont *correspondence: \hypersetup{pdfborder={0 0 0}}\href{mailto:charithc@cse.mrt.ac.lk}{\texttt{charithc@cse.mrt.ac.lk}}}}%
  }%
}

\usepackage{authblk}

\author[1]{Vipooshan Vipulananthan}
\author[1]{Kumudu Mohottala}
\author[1]{Kavindu Chinthana}
\author[1]{Nimsara Paramulla}
\author[1\thanks{\tt{charithc@cse.mrt.ac.lk}}]{Charith D. Chitraranjan}

\affil[1]{Department of Computer Science and Engineering, University of Moratuwa, Katubedda 10400, Sri Lanka (e-mail:
 \texttt{vipooshan.18@cse.mrt.ac.lk, kumudu.20@cse.mrt.ac.lk, kavinduc.20@cse.mrt.ac.lk, nimsara.20@cse.mrt.ac.lk, charithc@cse.mrt.ac.lk}}
\newcommand{\submittedtext}{%
\footnotesize\color{black}%
This is an open-access article published in IEEE Access. The final published version is available on IEEE Xplore at the following DOI:
\href{https://doi.org/10.1109/ACCESS.2025.3645127}{10.1109/ACCESS.2025.3645127}}

\newcommand{\topwatermark}{%
\begin{tikzpicture}[remember picture,overlay]
\node[anchor=north,yshift=-10pt] at (current page.north) {%
  \parbox{\dimexpr0.9\textwidth}{\submittedtext}%
};
\end{tikzpicture}%
}

\begin{document}

\twocolumn[ % Method A for two-column formatting
  \begin{@twocolumnfalse} % Method A for two-column formatting

\maketitle
% \submittednotice

% Place watermark at top of first page
\topwatermark

\begin{abstract}
Accident prediction and timely preventive actions improve road safety by reducing the risk of injury to road users and minimizing property damage. Hence, they are critical components of advanced driver assistance systems (ADAS) and autonomous vehicles. While many existing systems depend on multiple sensors such as LiDAR, radar, and GPS, relying solely on dash-cam videos presents a more challenging, yet more cost-effective and easily deployable solution. In this work, we incorporate improved spatio-temporal features and aggregate them through a recurrent network to enhance state-of-the-art graph neural networks for predicting accidents from dash-cam videos. Experiments using three publicly available datasets (DAD, DoTA and DADA) show that our proposed STAGNet model achieves higher average precision and mean time-to-accident scores than previous methods, both when cross-validated on a given dataset and when trained and tested on different datasets.
\end{abstract}
 % (optional)
\vspace{0.35cm}

  \end{@twocolumnfalse} % Method A for two-column formatting
] % Method A for two-column formatting

%\begin{multicols}{2} % Method B for two-column formatting (doesn't play well with line numbers), comment out if using method A

%%%%%%%%%%%%%%%  Main text   %%%%%%%%%%%%%%%
% \linenumbers

\section{Introduction}
\label{sec:introduction}
Accident prediction and timely warnings are vital for improving road safety by reducing injuries, fatalities, and property damage. Advanced driver assistance systems (ADAS) support human drivers by identifying potential hazards and are particularly valuable when they can anticipate accidents before they occur. Furthermore, accident anticipation provides an additional safety layer in autonomous vehicles~\cite{malawade2022spatiotemporal,fang2023vision, kumamoto2025aat}. While many existing systems depend on multiple sensors such as LiDAR, radar, and GPS, relying solely on dash-cam video offers a more cost-effective and easily deployable alternative. In this work, we address the problem of early accident anticipation using only dash-cam footage. We examine the limitations of existing methods and propose strategies to address them, thereby advancing vision-based accident anticipation.

Analyzing dash-cam video for driver assistance applications is a heavily researched area~\cite{fang2023vision, chitraranjan2025vision}. Among the many methods to anticipate future accidents, graph neural network (GNN)-based methods have demonstrated superior performance~\cite{thakur2024graph, mahmood2023new, malawade2022spatiotemporal}. In previous GNN-based methods~\cite{mahmood2023new, malawade2022spatiotemporal, bao2020uncertainty}, graphs are built from individual video frames, and temporal dependencies are modeled later with some form of a recurrent neural network (RNN). In recent work, Takur et al.~\cite{thakur2024graph} embed temporal dependencies into the construction of the graph itself by adding edges from nodes that correspond to past frames. They also introduce graph attention to focus on important frames. However, their approach does not directly model sequential information. In this work, we introduce a Long Short-Term Memory (LSTM) network to explicitly capture the temporal dependencies in the sequence of frames and embed them into the graph’s nodes. The key distinction of our approach is the integration of explicit sequence modeling with spatio-temporal graph attention. Furthermore, the feature extractors used in existing methods either ignore temporal dependencies, as they operate solely at the individual frame level (VGG16~\cite{simonyan2014very}), or are not tailored to videos with fast-moving objects (I3D~\cite{carreira2017quo}). We use SlowFast networks~\cite{Feichtenhofer2018SlowFastNF} to extract frame-level spatio-temporal features from sliding windows of the video stream. These networks capture both rapid motion dynamics (fast pathway) and slow-evolving spatial semantics (slow pathway). However, since SlowFast relies on temporal convolutions with small receptive fields (5 frames wide), it primarily captures short-term motion cues and is not designed to model long-term temporal dependencies. Introducing an LSTM module enables long-term temporal modeling. Similar ideas of augmenting temporal convolutions with LSTMs have produced improved performance in other domains~\cite{bi2021hybrid, hsu2022temporal}.

%based on their pre-trained weights (from human activity-related datasets), they are unable to learn features specific to road accidents, as their weights are kept frozen. Therefore, having the proposed LSTM module is important for learning accident-related temporal cues.

Another critical problem lies in the data used to experimentally evaluate existing systems. Some video datasets widely used to train and test accident anticipation methods have video quality bias (e.g., CCD~\cite{bao2020uncertainty}, A3D~\cite{yao2019unsupervised}) or a low proportion of ego-involved accidents (e.g., DAD~\cite{chan2017anticipating})~\cite{mahmood2023new}. However, evaluating these algorithms on unbiased datasets with a large proportion of ego-involved accidents is crucial, as such incidents are the ones that an ADAS can directly mitigate, rather than those solely involving other vehicles. Moreover, the generalization of models to datasets beyond those used to train them remains inadequately evaluated due to the use of biased datasets~\cite{chitraranjan2025vision}.

In summary, our contributions aimed at addressing the aforementioned problems are as follows.
\begin{itemize}
    \item  Encoding the input video frames with advanced spatio-temporal features capable of capturing rapid motion.
    \item Combining sequence modeling with graph attention by employing an LSTM that aggregates spatio-temporal features across frames before passing them on to the GNN.
    \item Applying graph attention to the object-level graphs.% in addition to the frame-level graphs.
    \item Providing a reduced version of the model for faster inference.
    \item Evaluating the proposed algorithm on unbiased datasets containing ego-involved collisions, including an analysis of its generalization capability across different datasets.
\end{itemize}

These architectural and feature extraction choices are simple, yet prove highly effective in practice and enable our model to outperform the state-of-the-art. We also show that the reduced version of our network is considerably faster than existing methods while still achieving higher average precision and time-to-accident values in predicting ego-involved accidents.

\section{Literature Review}

Typically, accident anticipation methods rely on pretrained convolutional neural networks (CNNs) to detect objects in each video frame and extract features from the detected objects and the full frame. We will refer to these as object-level and frame-level features, respectively. 

In one of the early works, Chan et al.~\cite{chan2017anticipating} proposed a method that feeds the extracted features through a spatial attention mechanism and a recurrent neural network (RNN). This allows the model to emphasize important objects based on cues such as appearance, location, and motion. They used VGG16~\cite{simonyan2014very} to extract object-level features. For frame-level features, they used VGG16 to extract features from each individual frame and IDT~\cite{wang2013action} to extract motion-related features from clips of 5 consecutive frames. Suzuki et al.~\cite{suzuki2018anticipating} proposed a similar method, with the main differences being the use of DeCAF~\cite{donahue2014decaf} features instead of VGG and a quasi-recurrent neural network~\cite{bradbury2016quasi} for the RNN. Another approach similar to~\cite{chan2017anticipating} was proposed by Karim et al.~\cite{karim2022dynamic}. They use a similar attention mechanism to weight objects in the same frame (spatial attention). In addition, however, they use a temporal attention scheme where the last $M$ hidden states of the RNN are aggregated as a weighted sum. This allows the model to focus more on important frames in the video.

Several authors have proposed graph neural network-based accident anticipation models. In one of the pioneering works, Bao et al.~\cite{bao2020uncertainty} construct a graph for each video frame such that the nodes represent the objects detected in the frame, and the edges represent the distances between objects in pixel space. Initial node features are defined as a concatenation of object-level and frame-level features extracted using a pretrained VGG16 model. The graph is then updated through a graph convolutional network (GCN) and a graph convolutional recurrent network (GCRN)~\cite{seo2018structured} before passing through a Bayesian neural network (BNN)~\cite{neal2012bayesian} that produces the probability of an accident and its associated uncertainty. Mahmood et al.~\cite{mahmood2023new} used a similar graph neural network architecture but observed that reliance on visual object features may reduce cross-dataset generalization and proposed replacing them with geometric features derived from bounding boxes, resulting in improved cross-dataset performance. Thakur et al.~\cite{thakur2024graph} proposed a nested graph approach that further improved performance. Their work forms the basis for our proposed method, and we describe it in more detail in the next section. 

The above studies summarize the methods most relevant to our approach. We refer readers to~\cite{fang2023vision, chitraranjan2025vision} for a more comprehensive review of accident anticipation literature.

\section{Methodology}

In this work, we propose an improved model to anticipate accidents from dashcam videos based on the Graph(Graph) framework~\cite{thakur2024graph}. Given a sequence of observed video frames \((X_1, \dots,X_{t-1}, X_t)\), the goal is twofold: (1) to predict whether an accident will occur or not, and (2) to detect it as early as possible. For each frame \(X_t\), the model outputs a probability \(p_t\) indicating the likelihood of an accident occurring in the future. For accident-positive cases, the Time-to-Accident (TTA) is defined as \(\tau = a - t\) for \(t < a\), where \(a\) is the start time of the accident and \(t\) is the earliest frame such that \(p_t > \alpha\), with \(\alpha\) being a decision threshold.

The architecture of our model is illustrated in Figure~\ref{fig1}. Similar to the Graph(Graph) model~\cite{thakur2024graph}, our model consists of three main modules. Module A captures the spatio-temporal relationships between detected objects, while module B extracts the global context of the video. Module C combines information from the other two modules to provide the final prediction. In the rest of this section, we describe these modules, highlighting the changes we have made with respect to the Graph(Graph) model.  
\subsection{Spatio-Temporal Object Graph Learning}

Not all objects captured by a dash-cam contribute equally to collision risk. Objects that are closer to each other and/or to the ego-vehicle pose a higher risk of collision. Although precise distance estimation is difficult with uncalibrated cameras, this module attempts to capture the relationships between traffic objects, both within each frame (spatial) and across adjacent frames (temporal). The model learns to focus more on relevant objects through the use of spatio-temporal graph attention mechanisms.

We use YOLO~\cite{wang2024yolov10realtimeendtoendobject} to detect objects and extract their bounding boxes from each video frame. For each bounding box, we extract visual features using a pretrained VGG16~\cite{simonyan2014very} network and the word embedding of the objects' class label using GloVe~\cite{pennington2014glove}. For a given frame \(X_t\), a spatio-temporal object graph \(G^{\text{obj}}_t\) with \(S\) nodes is constructed, where each node corresponds to a detected object, and $S$ is the number of objects detected. The features of each node are initialized with a concatenation of the corresponding objects' visual features \(f^{\text{obj}}_e \in \mathbb{R}^{S \times d_1}\) and the word embedding of its class label \(f^{\text{obj}}_l \in \mathbb{R}^{S \times d_2}\), where \(d_1\) and \(d_2\) are the visual feature and label embedding dimensions, respectively.

To model intra-frame spatial relationships, a spatial adjacency matrix \(A^{\text{obj}}_s\) is defined as follows. 
\begin{equation}
A^{\text{obj}}_s(i, j) = \frac{e^{-d(c_i, c_j)}}{\sum_{ij} e^{-d(c_i, c_j)}} \tag{1},
\end{equation}
where \(d(c_i, c_j)\) is the Euclidean distance (in pixel space) between the centers \(c_i\) and \(c_j\) of the \(i\)-th and \(j\)-th object bounding boxes detected in a frame, respectively. This ensures that spatially proximate objects contribute more, and distant or irrelevant objects are down-weighted during message passing in the Graph Attention Network that follows.

To model the movement of objects across time, we use a temporal adjacency matrix \(A^{\text{obj}}_{tm}\), which links objects of the same class across frames based on the cosine similarity of their features. This temporal graph structure allows the model to propagate information between corresponding objects over time. The temporal adjacency matrix $A^{\text{obj}}_{tm}(i, j)$ is defined as:

\begin{equation}
A^{\text{obj}}_{tm}(i, j) = 
\begin{cases} 
s(i, j), & \text{if } l_i = l_j \\
0, & \text{otherwise}
\end{cases}, \tag{2}
\end{equation}
where \(s(i, j)\) is the cosine similarity between the object features of the \(i\)-th and \(j\)-th objects with labels \(l_i\) and \(l_j\), belonging to frames at time \(t\) and \(t - u\), respectively. These edges propagate information between objects of the same class across frames and avoid irrelevant information flow between objects from different classes.

The node embeddings are updated through Graph Attention Networks (GATv2)~\cite{brody2021attentive} using the following operations:

\setcounter{equation}{2}  % Set equation number to 2 so that next one will be 3
\begin{equation}
\begin{aligned}
f^{s}_{\text{obj}} &= \text{GAT}([\phi(f^{\text{obj}}_e), \phi(f^{\text{obj}}_l)], A^{\text{obj}}_s) \\
f^{tm}_{\text{obj}} &= \text{GAT}([\phi(f^{\text{obj}}_e), \phi(f^{\text{obj}}_l)], A^{\text{obj}}_{tm}) \\
f'_{\text{obj}} &= [f^{s}_{\text{obj}}, f^{tm}_{\text{obj}}],
\end{aligned}
\end{equation}

where \(\phi\) denotes a fully connected layer for dimensionality reduction, and \([\cdot]\) represents concatenation. Note that the original Graph(Graph) architecture used Graph Convolutional Networks (GCN) for this computation. We adopt GATv2 instead of GCNs because its attention mechanism allows each node to attend selectively to the most relevant neighbors. This graph attention approach, particularly its improved version GATv2, has shown better performance than non-attentive methods such as GCNs on some of the benchmark graph datasets~\cite{brody2021attentive}. This attention mechanism can help our model learn to emphasize the most relevant object relationships in the sequence of scenes, improving prediction accuracy and supporting early accident anticipation.

\subsection{Spatio-Temporal Frame-level Feature Learning}

While the spatio-temporal object graph captures fine-grained object interactions, extracting global contextual features from the entire frame sequence is crucial for providing a holistic understanding of the scene. Most previous approaches~\cite{bao2020uncertainty, mahmood2023new, karim2022dynamic} rely on pretrained image models such as VGG16~\cite{simonyan2014very} for global feature extraction. However, this does not capture temporal features that span across frames. Thakur et al.~\cite{thakur2024graph} extract global frame-level features with I3D~\cite{carreira2017quo}, which derives features from a sequence of frames, thus capturing both spatial and temporal features. In this work, we utilize a pretrained SlowFast network~\cite{Feichtenhofer2018SlowFastNF} to extract frame-level spatio-temporal features. SlowFast networks have demonstrated better performance than I3D on video recognition benchmarks and have a dedicated path to process fast movements efficiently. This is useful in processing dash-cam videos, which often contain fast-moving objects.

To extract the global feature representation \( f_{\text{fr}}^t \) for a frame at time \( t \), we pass the current frame and the last-$u$ frames \( X_{\text{seq}} = (X_{t-u}, \dots, X_t) \) through the SlowFast network, which produces a high-dimensional spatio-temporal representation. These features are then projected into a lower-dimensional space using a fully connected (FC) layer. To further enhance temporal modeling, we pass the transformed features through a Long Short-Term Memory (LSTM) module:

\setcounter{equation}{3}  % Adjust as needed
\begin{equation}
\begin{aligned}
f_{\text{fr}} &= SlowFast(X_{\text{seq}}) \\
f^{\text{fc}}_{\text{fr}} &= \phi(f_{\text{fr}}) \\
f'_{\text{fr}} &= \text{LSTM}(f^{\text{fc}}_{\text{fr}}),
\end{aligned}
\end{equation}

where $\phi$ denotes a fully connected layer used for dimensionality reduction. This enriched global representation is subsequently used in the higher-level Spatio-Temporal Global Feature Learning module to support accurate accident anticipation.

\subsection{Frame Graph Learning}

To capture temporal relationships between video frames, a \textit{frame graph} is constructed that leverages both global frame-level features and object-level features obtained from the previous stage. In this graph, each node corresponds to a frame at a particular time step \(t\), and edges are directed from past frames to the current one, ensuring that no future information is leaked backward in time. Each frame node is connected to its previous \(k\) frames through \(k\) directed edges. The adjacency matrix \(A_{\text{fr}}\) for the frame graph is defined as:

\begin{equation}
A_{\text{fr}}(i, j) =
\begin{cases}
1, & \text{if } i - j \leq k \\
0, & \text{otherwise}
\end{cases},
\end{equation}

where \(i\) is the index of the current frame node.

Two parallel Graph Attention Network (GAT) layers are applied to this graph. The first GAT layer processes the object-level features \(f'_{\text{obj}}\), and the second processes the global frame-level features \(f'_{\text{fr}}\):

\begin{equation}
\begin{aligned}
f''_{\text{obj}} &= \text{GAT}(f'_{\text{obj}}, A_{\text{fr}}) \\
f''_{\text{fr}} &= \text{GAT}(f'_{\text{fr}}, A_{\text{fr}}) \\
f'' &= [f''_{\text{obj}}, f''_{\text{fr}}]
\end{aligned}
\end{equation}

The purpose of applying attention here is to learn which frames are more significant for predicting upcoming accidents. The concatenated feature vector \(f''\) is then passed through fully connected layers to compute the frame-level accident probabilities \((p_1, p_2, \dots, p_N)\).

The entire model is trained end-to-end, except for the object detection and global feature extraction modules, which are kept frozen. Standard cross-entropy loss is used for optimization:

\begin{equation}
\mathcal{L}(p, y) = - \sum_{m=1}^{M} y_m \log \left( \frac{e^{p_m}}{\sum_j e^{p_j}} \right),
\end{equation}

where \(M\) is the number of classes (2 in this case), \(p\)'s denote the predicted logits, and \(y_m=1\) for the correct ground-truth class label and \(y_m=0\) for all other classes, for a given frame.

\begin{figure*}[t!] % [!t] forces top placement aggressively
    \centering
    \includegraphics[width=\textwidth, height=0.75\textheight, keepaspectratio]{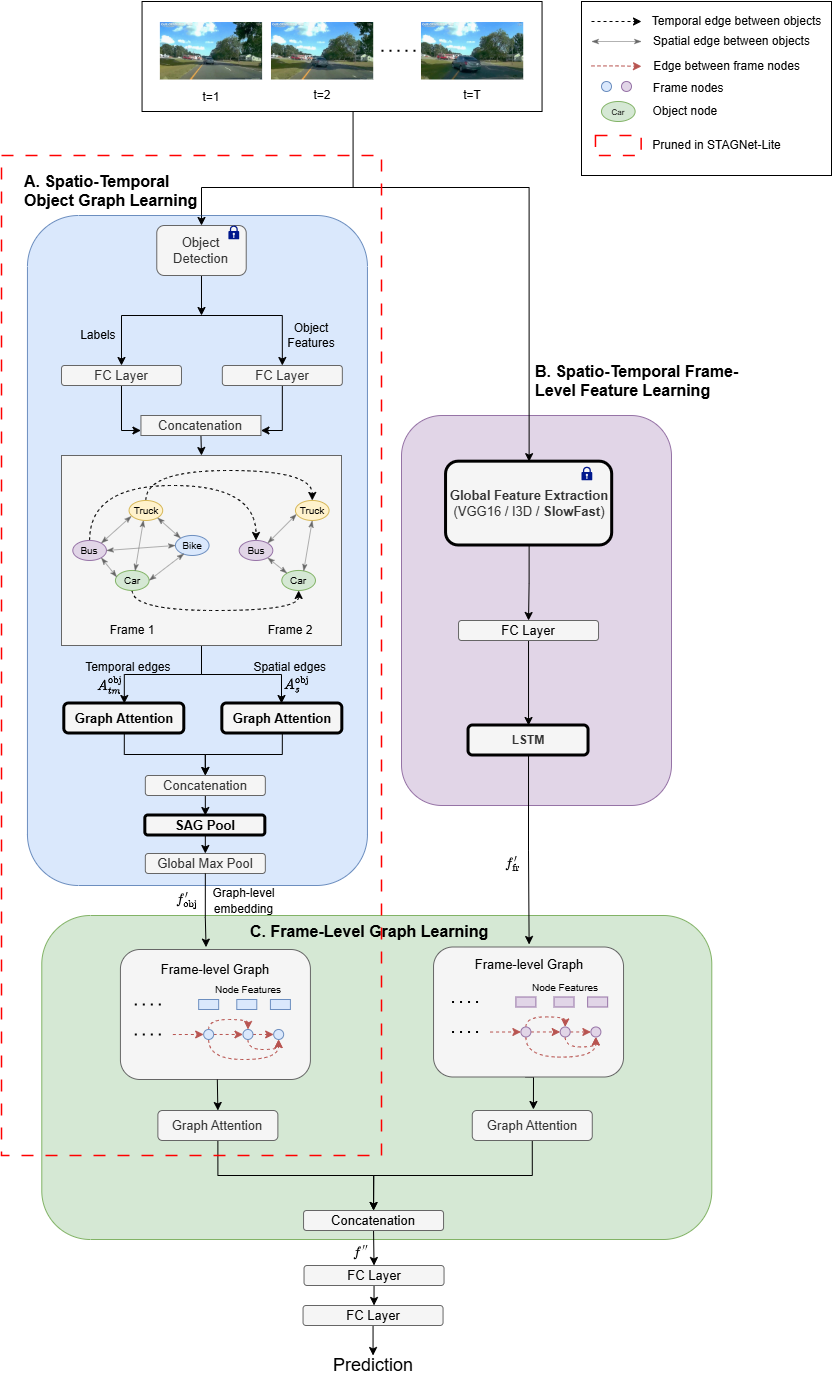}
    \caption{Architecture of the proposed STAGNet accident anticipation framework. Boxes with thicker borders represent the main components introduced in this work. The components within the red dashed rectangle are pruned in the light-weight model, STAGNet-Lite.}
    \label{fig1}
\end{figure*}

\subsection{Model Pruning}

We note that extracting object-level features from each bounding box and constructing object graphs is computationally expensive and may be unnecessary for ego-involved accident anticipation. Since the ego-vehicle is not detected as an object, its relationship to other objects is not modeled in the object graphs. Hence, we hypothesized that the spatio-temporal \textit{frame-level} feature learning module and its corresponding frame-level graph are sufficient to predict ego-involved accidents. Therefore, we propose a lightweight version of our model (STAGNet-Lite) by pruning the spatio-temporal \textit{object} graph learning module and its corresponding frame-level graph for the efficient anticipation of ego-involved accidents. The red dashed rectangle in Figure~\ref{fig1} encapsulates the pruned components.

\section{Experimental Evaluation}
In this section, we first present the details of the experimental setup, datasets used, the baseline methods, and the evaluation metrics. We then discuss the outcomes of the experiments conducted.

\subsection{Experimental Settings}

\subsubsection{Datasets}

We conducted experiments on three benchmark dash-cam video datasets: DAD~\cite{chan2017anticipating}, DoTA~\cite{yao2022dota}, and DADA~\cite{fang2019dada}. 

The Dashcam Accident Dataset (DAD) consists of 1,750 five-second clips derived from 678 dashcam videos recorded at 20 fps. Among them, 620 clips are annotated as ‘accident’ (positive) and 1,130 as ‘normal’ (negative). In all positive clips, the accident event occurs within the final 10 frames. Note that this dataset primarily consists of non-ego-involved accidents (less than 10\% ego-involved incidents)~\cite{karim2022dynamic}. We used the standard train-test split provided by the authors of Graph(Graph) \footnote{\url{https://github.com/thakurnupur/Graph-Graph}}, 
which is the same as that used in other work as well. This ensures a fair and consistent comparison with prior work. 

Since the anticipation of ego-involved accidents is more important, we evaluated our model on two widely used accident datasets containing a large percentage of ego-involved accidents: DoTA~\cite{yao2022dota} and DADA~\cite{fang2019dada}. 

From the DADA dataset, we selected all clips labeled as \textit{ego-involved accidents} and extracted five-second segments such that the accident onset occurred within the last two seconds, with the exact start time randomized (as in CCD). This produced 938 positive samples. Negative samples were obtained by extracting five-second segments from normal driving periods preceding the accident windows. Although we considered each video in the dataset as a potential source for a negative sample, those lacking a sufficiently long pre-accident segment were excluded. This procedure yielded 1361 negative samples.

Positive and negative clips from the DoTA dataset were extracted following a similar procedure. Based on the duration of the video clips and the accident onset times, we were able to extract 745 negative samples depicting normal driving scenes. To construct a balanced dataset\footnote{A balanced dataset was used to prevent inflated AP scores that can occur when positive samples greatly outnumber negative ones~\cite{saito2015precision}.}, we randomly selected an equal number of videos containing ego-involved accidents to produce 745 positive samples. Note that Mahmood et al.~\cite{mahmood2023new} have followed a similar approach to select videos from DoTA and reported similar numbers of videos. The decision to set the duration of each extracted clip to five seconds was based on other popular datasets, e.g., DAD and CCD.

We adopt a $k$-fold cross-validation strategy for DoTA and DADA, which provides a more robust evaluation of model performance. During cross-validation, we group clips so that any samples from the same source video are in the same fold. 

A summary of the characteristics of these datasets is presented in Table~\ref{tab:dataset_stats}, and more details can be found in~\cite{chitraranjan2025vision}. The videos and the relevant scripts can be found \href{https://github.com/Vipooshan1998/STAGNet}{at this link}. 

These datasets provide a diverse set of challenging driving scenarios that are essential for testing early accident anticipation performance in real-world conditions. To further evaluate model generalization, we conducted a cross-dataset evaluation, where the model was trained on DoTA and tested on DADA, and vise versa. Note that we down-sampled the videos in DADA from 30 fps to 10 fps for compatibility with those in DoTA to enable direct comparison of results. 

Note that we did not include the popular dataset CCD~\cite{bao2020uncertainty} in our evaluation due to its quality bias, caused by the difference in video quality between accident and normal videos~\cite{mahmood2023new}. This bias can be exploited by models to classify videos based on quality rather than on accident cues. This is evident in the near-perfect performance of all models on CCD.

\begin{table}[htbp]
\scriptsize
\caption{Dataset Statistics.}
\centering
\begin{tabular}{|l|c|c|c|c|}
\hline
\textbf{Dataset} & \textbf{Positive} & \textbf{Negative} & \textbf{Total Videos} & \textbf{FPS} \\
\hline
DAD~\cite{chan2017anticipating} & 620 & 1130 & 1750 & 20 \\
DoTA~\cite{yao2022dota} & 745 & 745 & 1490 & 10 \\
DADA~\cite{fang2019dada} & 938 & 1361 & 2299 & 10 \\
\hline
\end{tabular}
\label{tab:dataset_stats}
\end{table}

\subsubsection{Networks and Baselines}

\begin{figure}[h]
    \centering
    \includegraphics[width=0.9\linewidth]{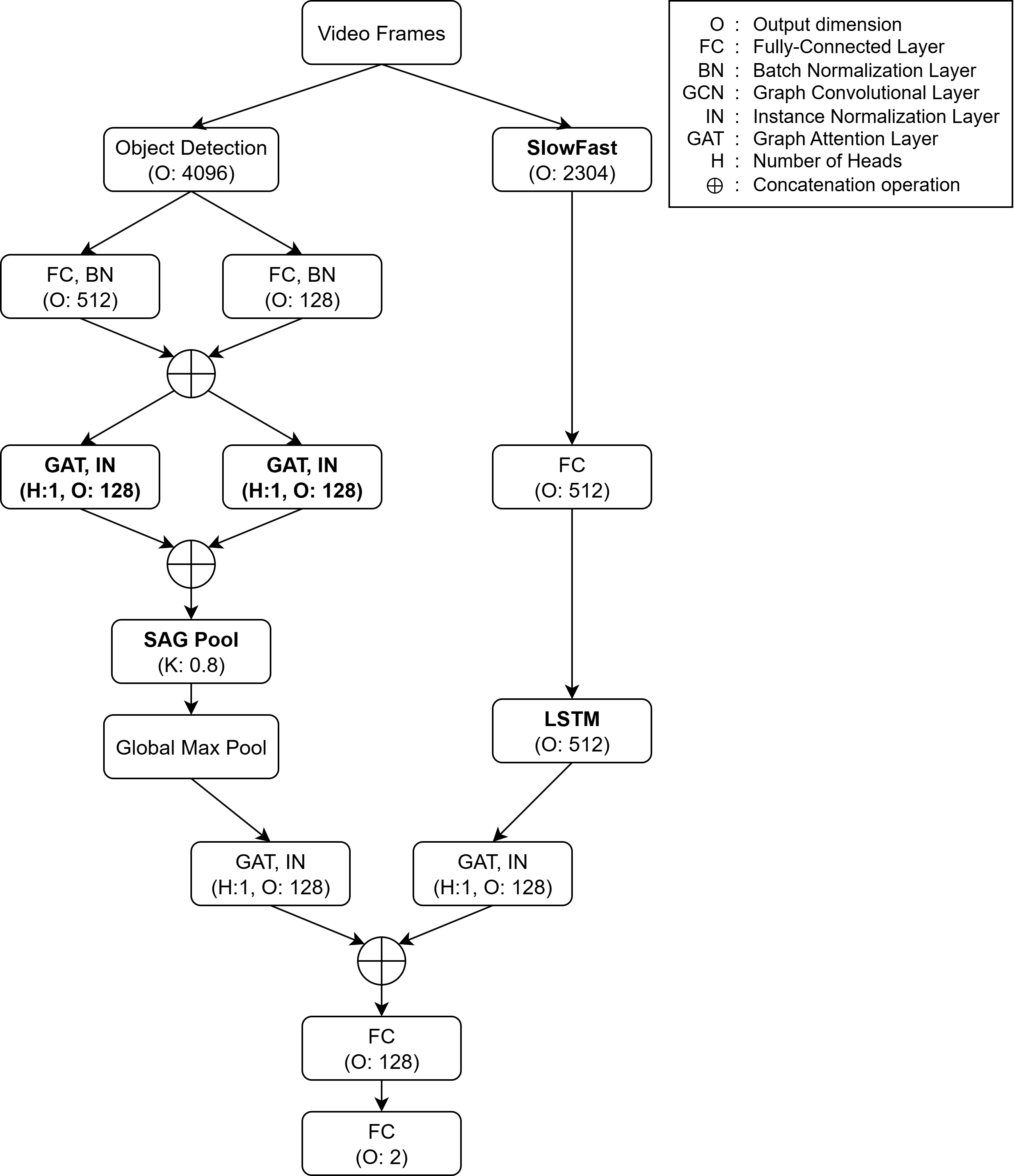}
    \caption{Configuration of the proposed STAGNet accident anticipation framework.}
    \label{fig:architecture}
\end{figure}

We use the pretrained YOLO~\cite{wang2024yolov10realtimeendtoendobject} and VGG16~\cite{simonyan2014very} models to detect objects and extract object-level features, respectively. Specifically, we utilize the top-19 objects detected, each represented by a feature vector of dimension $d_1 = 4096$. To encode object labels, we apply GloVe word embeddings \cite{pennington2014glove} using the spaCy library, resulting in embedding vectors of size $d_2 = 300$. When setting temporal edges in the object graphs, each object node is connected only to its counterpart in the previous frame ($u = 1$). For frame-level graphs, we set the number of temporal neighbors to $k = 20$~\cite{thakur2024graph}. The same architecture and configuration, illustrated in Figure~\ref{fig:architecture}, are used across all datasets.

For global visual feature extraction, we adopt the SlowFast network~\cite{Feichtenhofer2018SlowFastNF}, which produces a $h$-dimensional feature vector for each frame, where $h = 2304$. All experiments are implemented using the PyTorch framework. 

For our re-implementation of AAT-DA~\cite{kumamoto2025aat}, we used the model parameters presented in the original paper, as follows. The same feature extraction pipeline that is commonly used in previous methods~\cite{bao2020uncertainty, karim2022dynamic} was used: Top-19 objects detected were considered, and 4096-dimensional object-level and global features were extracted using VGG16. For all the transformers, the hidden layer dimension and the number of attention heads were set to 1024 and 8, respectively. Dropout rates of 0.3, 0.3, 0.1, and 0.5 were applied to the spatial transformer, object self-attention layer, temporal transformer, and the fully connected layer, respectively.

On the DAD dataset, we compare our approach against several accident anticipation methods: DSA \cite{chan2017anticipating}, adaLEA \cite{suzuki2018anticipating}, UString \cite{bao2020uncertainty}, DSTA~\cite{karim2022dynamic}, EGSTL~\cite{mahmood2023new}, AAT-DA~\cite{kumamoto2025aat}, and Graph(Graph)~\cite{thakur2024graph}. These algorithms were selected based on their relevance to this research and widespread use as baselines in similar work. The results for these methods are directly taken from the respective publications. The results for our proposed STAGNet model are based on the same training/testing splits of DAD as those used by these comparison methods.  

On the DoTA and DADA datasets, we compared our model against UString \cite{bao2020uncertainty}, UString (Geometric), Graph(Graph)~\cite{thakur2024graph}, and the recent transformer-based model AAT-DA~\cite{kumamoto2025aat}. We were unable to compare with the remaining methods because their implementations were not available for us to run on the DoTA and DADA datasets. Note that we implemented UString (Geometric) as an attempt to approximate GSTL~\cite{mahmood2023new} by replacing the VGG16-based object-level features in UString \cite{bao2020uncertainty} with geometric features derived from the detected bounding boxes. Furthermore, although Mahmood et al.~\cite{mahmood2023new} used a similar number of videos from the DoTA dataset, their selected subset of accident videos differs from ours. We specifically selected ego-involved accidents, whereas they did not. Thus, the DoTA results reported in~\cite{mahmood2023new} are not comparable to those in our study.

Our STAGNet model was implemented in PyTorch version 2.0.0 and trained for 20 Epochs using the Adam optimizer with a batch size of 1 and a learning rate of 0.0001. These same hyper-parameters, which are inherited from~\cite{bao2020uncertainty, thakur2024graph}, were applied to all three datasets.

\subsubsection{Evaluation Metrics}

Accident anticipation models are expected to be both \textit{accurate} and \textit{early} in their predictions. Consistent with prior work \cite{bao2020uncertainty, mahmood2023new,thakur2024graph, fang2023vision}, we evaluate our model using \textit{Average Precision (AP)} and \textit{mean Time-to-Accident (mTTA)}. AP reflects the correctness of predictions by measuring how well the model distinguishes between frames leading to an accident and those without an accident in the near future. For a given frame, if the predicted probability \( p_t \) of a future accident exceeds a threshold \( \alpha \), it is counted as a positive prediction. By varying \( \alpha \), a precision-recall curve is generated, and AP is computed as the area under this curve. A higher AP indicates more accurate predictions.

To determine how early the model can anticipate an accident, we use the \textit{Time-to-Accident (TTA)} metric, which measures the time difference between the actual onset of an accident and the earliest frame for which \( p_t > \alpha \). By evaluating TTA across different thresholds \( \alpha \), we compute the \textit{mean TTA (mTTA)} as the average of all corresponding values. %However, a high mTTA accompanied by low AP may reflect poor performance due to many false positives. Therefore, we report the mTTA value corresponding to the threshold that yields the highest AP.

%To reduce bias introduced by data splitting, we perform \textit{5-fold cross-validation}, and report the mean of the AP and mTTA values over all folds.

\subsection{Results and Discussion}

\subsubsection{Comparison with State-of-the-art Methods}

Table~\ref{tab:dad_results} shows the results of the experiments on DAD. Our proposed algorithm achieves the highest AP and mTTA, closely followed by Graph(Graph)~\cite{thakur2024graph}. These two methods perform considerably better than the other comparison methods. The transformer-based method, AAT-ODA~\cite{kumamoto2025aat}, achieves a very competitive AP but at the expense of a lower mTTA. 

\begin{table}[htbp]
\scriptsize
\caption{Performance comparison on the dash-cam accident dataset (DAD)}
\centering
\begin{tabular}{|c|c|c|c|}
\hline
\textbf{Dataset} & \textbf{Method} & \textbf{AP (\%)} & \textbf{mTTA (s)} \\
\hline
\multirow{5}{*}{DAD ~\cite{chan2017anticipating}} 
& DSA~\cite{chan2017anticipating}       & 48.10 & 1.34 \\
& adaLEA~\cite{suzuki2018anticipating}   & 52.30 & 3.43 \\
& UString~\cite{bao2020uncertainty}  & 53.70 & 3.53 \\
& DSTA~\cite{karim2022dynamic}  & 56.1 & 3.66 \\
& EGSTL~\cite{mahmood2023new}       & 56.30 & 3.23 \\
& Graph(Graph) ~\cite{thakur2024graph}       & 62.80 & 4.29 \\
& AAT-DA ~\cite{kumamoto2025aat}       & 64.0 & 2.87 \\
& STAGNet-Lite (Ours)        & 50.47 & 4.24 \\
& STAGNet (Ours)        & \textbf{64.36} & \textbf{4.32} \\
\hline
\end{tabular}
\label{tab:dad_results}
\end{table}

As shown in Table~\ref{tab:dota_dada_comparison}, for the DoTA and DADA datasets containing ego-involved accidents, our model consistently predicts accidents more precisely and earlier than the comparison methods. On DoTA, the UString (Geometrics) and UString models achieve average precisions of 63.16\% and 75.40\%, respectively, while the Graph(Graph) model reaches 83.50\%. We also evaluated the recent transformer-based method AAT-DA~\cite{kumamoto2025aat}, which achieved an AP of 72.45\%\footnote{Note that the implementation of AAT-DA was not made available by the authors. Therefore, we implemented it based on the details given in the paper to the best of our knowledge. For the DAD dataset, we were able to achieve very similar results (63\% AP) to those reported in their work (64\% AP), suggesting a reasonable re-implementation.}. In comparison, our STAGNet model achieves an average precision of 92.73\% and an mTTA of 3.23 seconds. Our pruned model, STAGNet-Lite, also achieves comparable results, with an average precision of 91.89\% and an mTTA of 3.13 seconds. On the DADA dataset, our model achieves 95.39\% AP, outperforming UString (Geometrics) at 56.75\%, UString at 63.95\%, AAT-DA at 69.36\%, and Graph(Graph) at 84.04\%. The corresponding mTTA of 3.08 seconds also confirms our methods' ability to anticipate accidents earlier. STAGNet-Lite achieves an average precision of 94.42\% and an mTTA of 3.11 seconds. %These consistent improvements across multiple datasets highlight the effectiveness and robustness of our proposed framework in capturing rich spatio-temporal patterns for early and accurate accident anticipation. In particular, our improved global feature pipeline has a marked effect on predicting ego-involved accidents. 

\begin{table}[htbp]
\caption{Performance comparison of methods on DoTA and DADA Datasets}
\centering
\scriptsize
\begin{tabular}{|c|c|c|c|}
\hline
\textbf{Dataset} & \textbf{Method} & \textbf{AP (\%)} & \textbf{mTTA (s)} \\
\hline
\multirow{4}{*}{DoTA ~\cite{yao2022dota}} & UString (Geometrics) ~\cite{mahmood2023new} & 63.16 & 2.14 \\
                    & UString ~\cite{bao2020uncertainty}                & 75.40 & 2.26 \\
                    & Graph(Graph)~\cite{thakur2024graph}               & 83.50 & 3.14 \\
                    & AAT-DA*~\cite{kumamoto2025aat}                     & 72.45 & \textbf{3.28} \\
                    & STAGNet-Lite (Ours) 
                            & 91.89 & 3.13 \\
                    & STAGNet (Ours)                 &                \textbf{92.73} & 3.23 \\
\hline
\multirow{4}{*}{DADA ~\cite{fang2019dada}} & UString (Geometrics) ~\cite{mahmood2023new} & 56.75 & 1.25 \\
                      & UString  ~\cite{bao2020uncertainty}            & 63.95 & 1.03 \\
                      & Graph(Graph) ~\cite{thakur2024graph}                & 84.04 & 2.92 \\
                    & AAT-DA*~\cite{kumamoto2025aat}                         & 69.36 & 1.36 \\
                    & STAGNet-Lite (Ours) 
                            & 94.42 & \textbf{3.11} \\
                    & STAGNet (Ours)                 & \textbf{95.39} & 3.08 \\
\hline
\end{tabular}
\label{tab:dota_dada_comparison}
\begin{flushleft}
\scriptsize \textit{*} Re-implemented based on the details given in the original paper. 
\end{flushleft}
\vspace{-4pt}
\end{table}

\subsubsection{Cross-dataset Evaluation}

To assess the ability of our model to generalize, we conducted cross-dataset experiments by training on one dataset and evaluating it on another. During this process, any hyperparameter tuning was confined to the training dataset. As shown in Table~\ref{tab:cross_dataset_generalization}, our method outperforms the comparison methods across both datasets. When trained on the DoTA dataset and tested on DADA, our method achieves an AP of 67.47\%, outperforming UString (Geometrics) at 46.39\%, UString at 48.26\%, AAT-DA at 50.27\%, and Graph(Graph) at 61.34\%. Additionally, it achieves a higher mTTA of 3.28 seconds, indicating better anticipation of accidents. Similarly, when trained on DADA and evaluated on DoTA, our method achieves an AP of 80.15\%, surpassing UString (Geometrics) at 54.85\%, UString at 55.63\%, AAT-DA at 66.23\%, and Graph at 68.55\%. It also records a higher mTTA of 3.13 seconds. These results highlight the better generalization ability of our model across datasets.

However, the cross-dataset performance of all tested algorithms is considerably worse than the cross-validation results on the same datasets. To investigate possible reasons for this drop in performance, we probed for differences in video quality and the distribution of object categories involved in the accidents between the DoTA and DADA datasets. 

\begin{figure}[!t]
    \centering
    \includegraphics[width=\columnwidth]{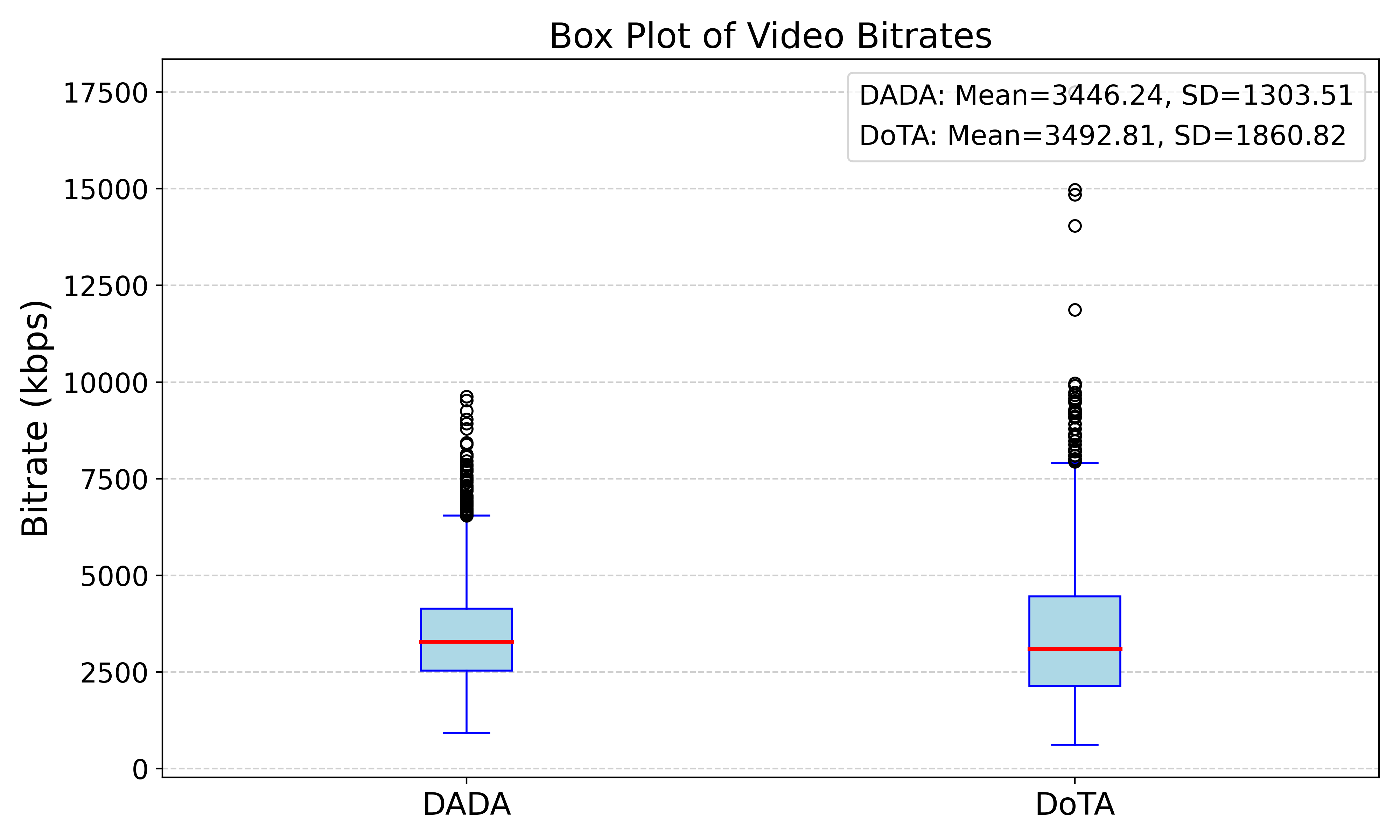}
    \caption{Box plots showing the bit-rates of DoTA and DADA videos.}
    \label{fig:bitrate_boxplot}
\end{figure}

\begin{figure*}[!t]
    \centering
    % ---------- Top row: DADA ----------
    \begin{subfigure}[b]{0.47\textwidth}
        \centering
        \includegraphics[width=\linewidth]{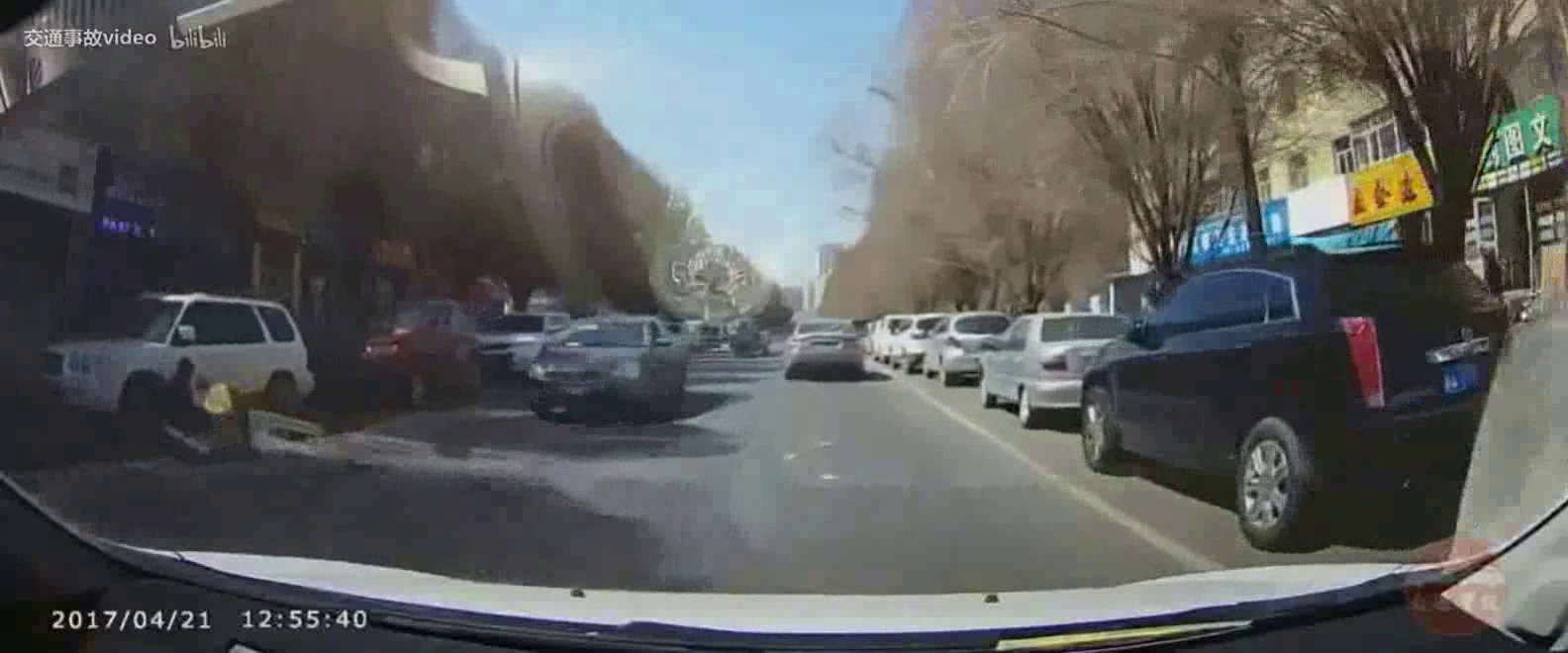}
        \label{fig:dada_sample_1}
    \end{subfigure}
    \hspace{0.015\textwidth} % reduced horizontal gap
    \begin{subfigure}[b]{0.47\textwidth}
        \centering
        \includegraphics[width=\linewidth]{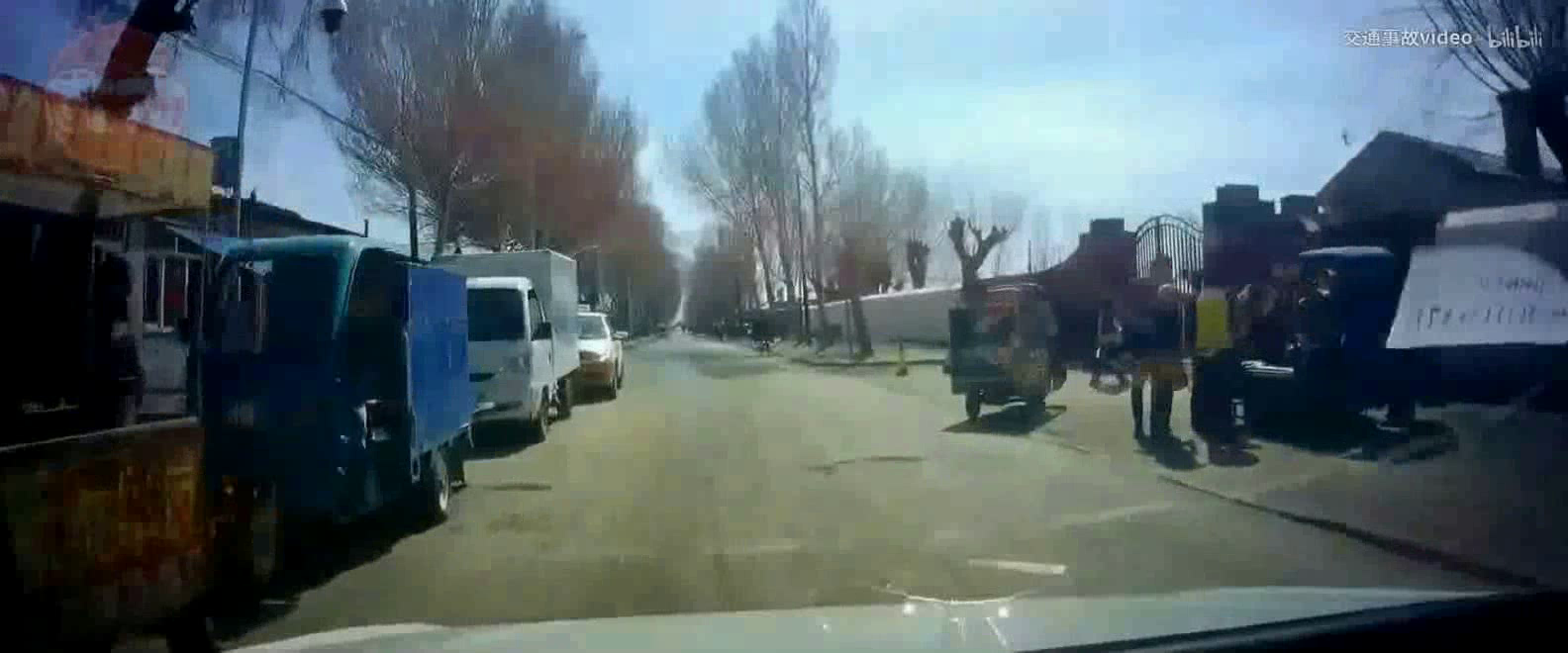}
        \label{fig:dada_sample_3}
    \end{subfigure}

    \vspace{0.5em} % reduced vertical gap between rows

    % ---------- Bottom row: DoTA ----------
    \begin{subfigure}[b]{0.47\textwidth}
        \centering
        \includegraphics[width=\linewidth]{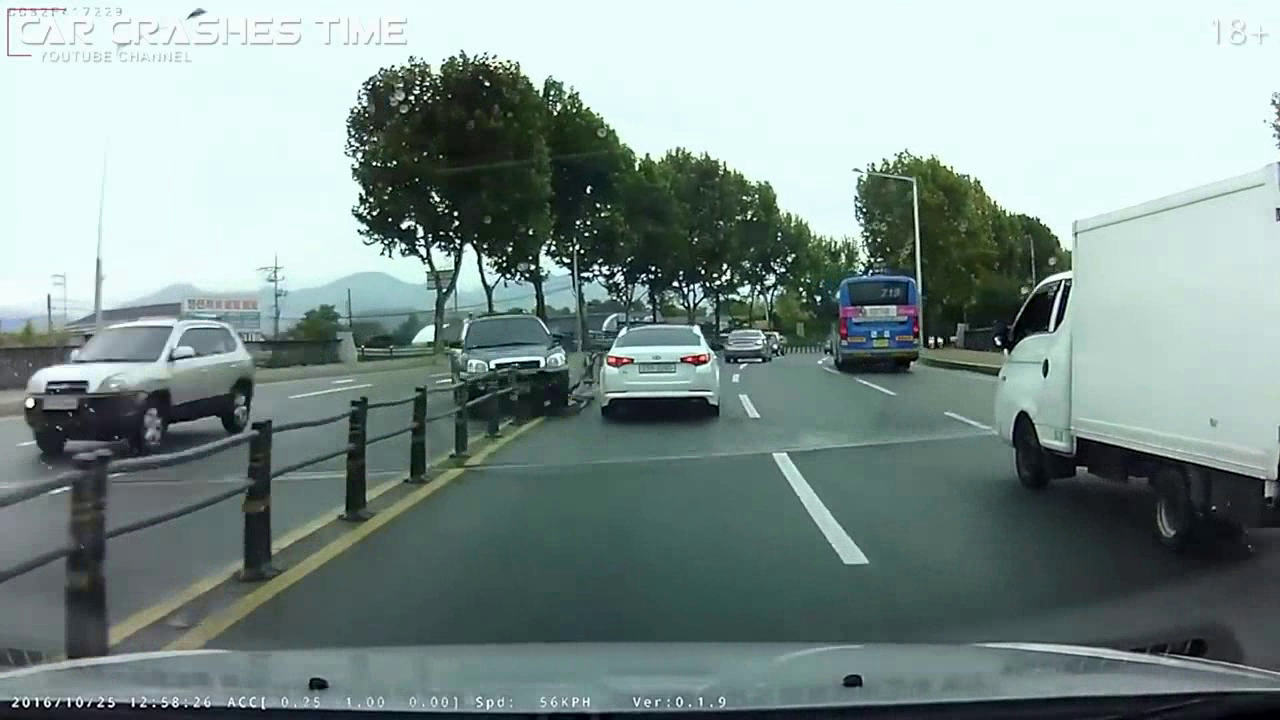}
        \label{fig:dota1}
    \end{subfigure}
    \hspace{0.015\textwidth} % reduced horizontal gap
    \begin{subfigure}[b]{0.47\textwidth}
        \centering
        \includegraphics[width=\linewidth]{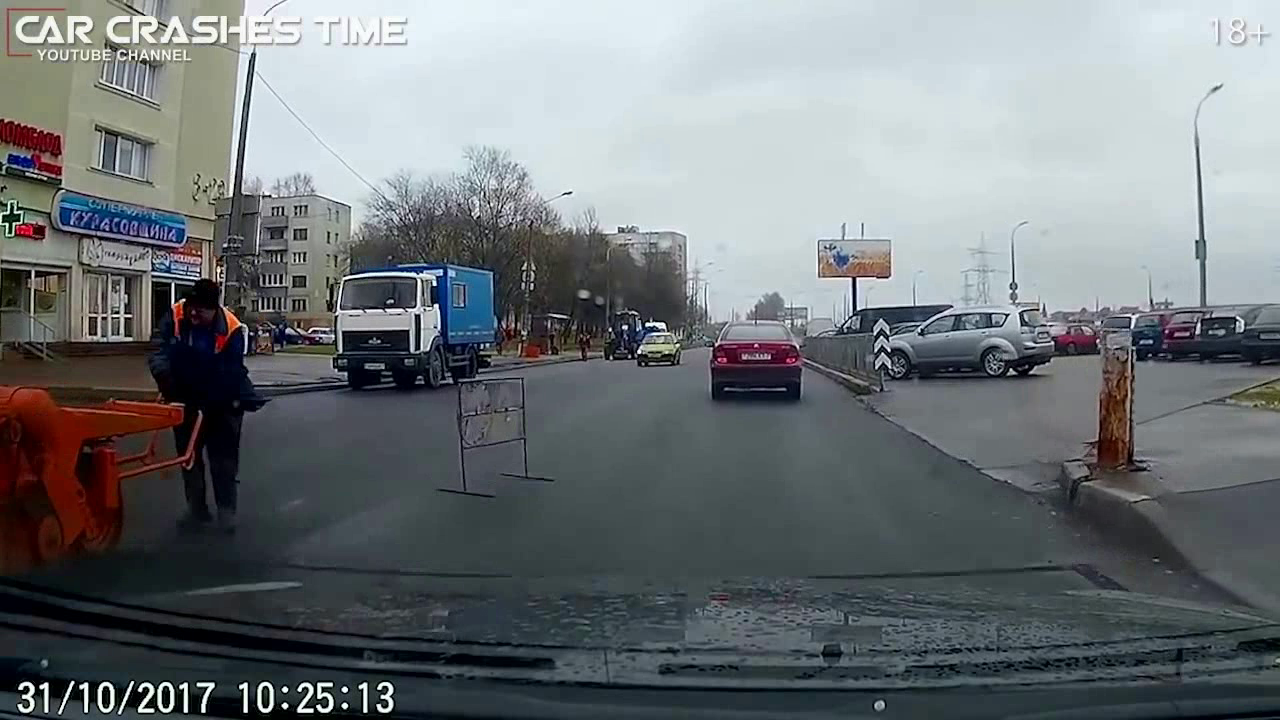}
        \label{fig:dota3}
    \end{subfigure}

    \caption{Sample frames from DADA (top row) and DoTA (bottom row). The visual quality of DADA videos is lower than that of DoTA videos.}
    \label{fig:dada_dota_samples}
\end{figure*}

Figure~\ref{fig:bitrate_boxplot} reveals that there is no significant difference between the bit-rates of DoTA and DADA videos. However, there is a noticeable difference in visual quality between the two datasets. Figure~\ref{fig:dada_dota_samples} shows a few samples. Videos in DADA use the MPEG-4 Visual Simple Profile (SP) codec, whereas those in DoTA use Xvid. Even at similar bit-rates, the Xvid codec may use MPEG-4 Advanced Simple Profile (ASP) features and produce higher quality videos\footnote{https://labs.xvid.com/project/}. This difference in quality may contribute to the drop in cross-dataset performance. In particular, the higher quality of DoTA videos can produce clearer object boundaries and smoother motion, which can enable the model to learn more detailed spatial-temporal features. However, these learned features may not generalize well to the lower-quality videos in DADA.

\begin{figure*}[!t]
    \centering
    \begin{subfigure}[b]{0.47\textwidth}
        \centering
        \includegraphics[width=\linewidth]{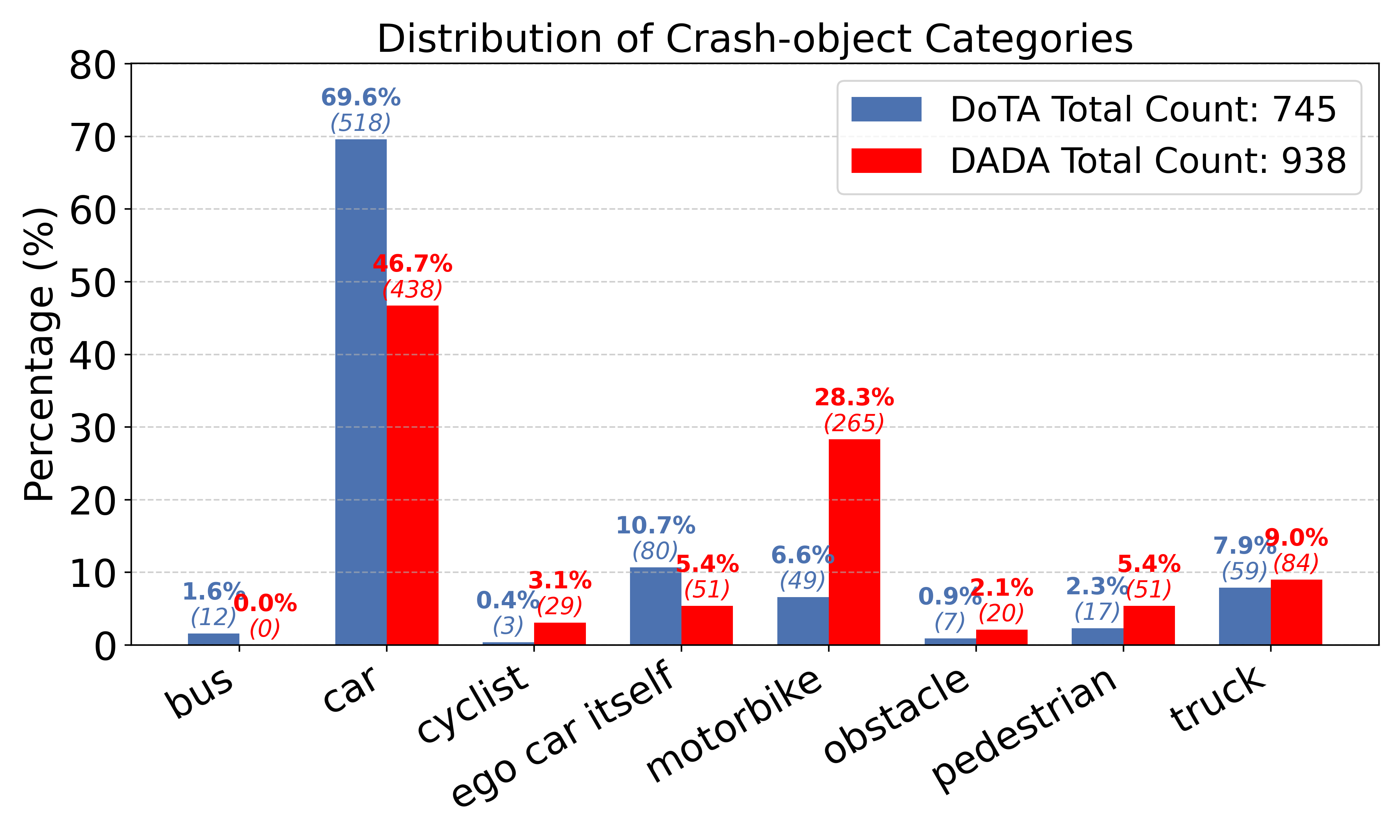}
        \label{fig:dada_co_dist}
        \caption{Distribution of crash-object categories.}
    \end{subfigure}
    \hspace{0.04\textwidth} % reduced horizontal gap
    \begin{subfigure}[b]{0.47\textwidth}
        \centering
        \includegraphics[width=\linewidth]{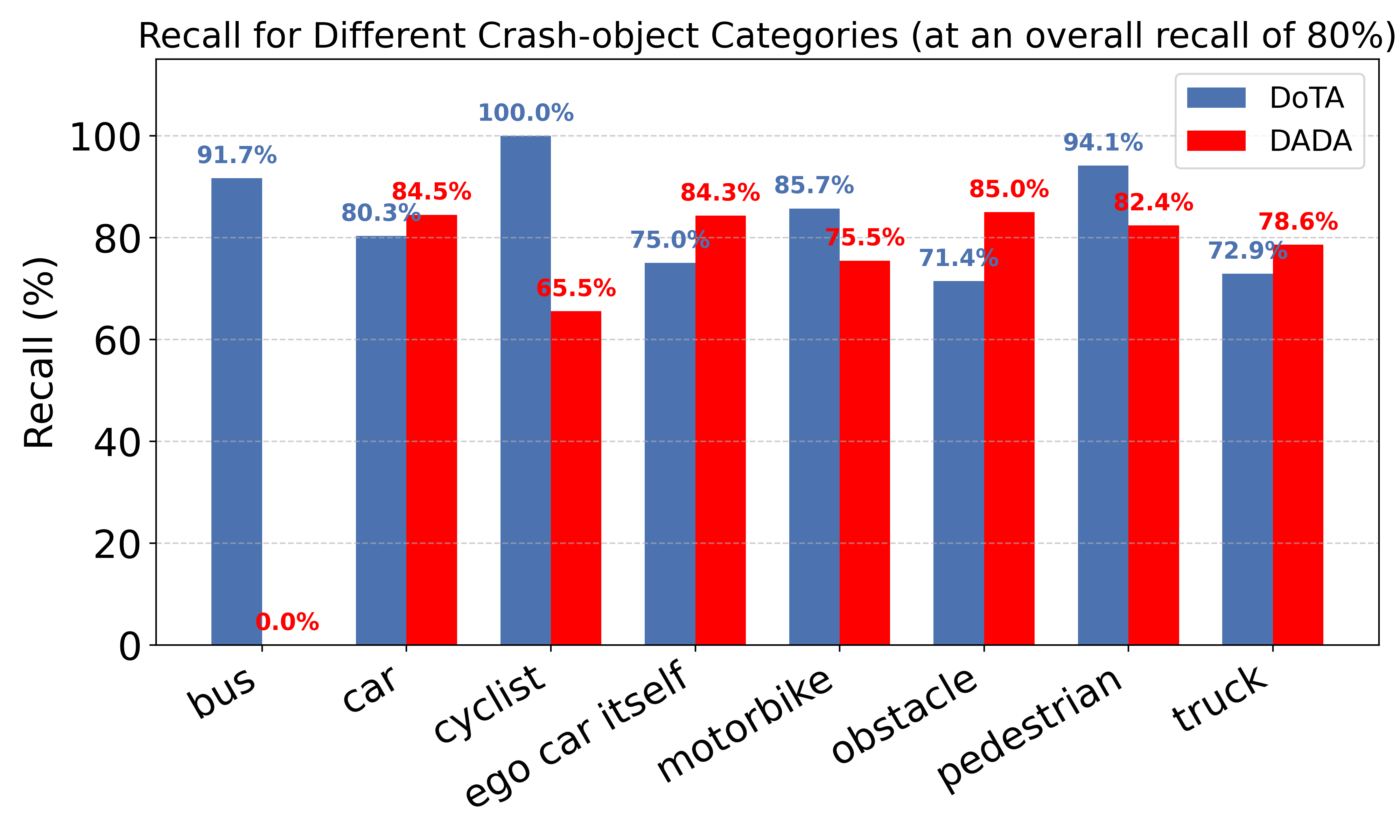}
        \label{fig:dada_co_recall}
        \caption{Cross-dataset recall for different crash-object categories (at an overall recall of 80\%).}
    \end{subfigure}
    % Main/common caption
    \caption{(a) Distribution of crash-object categories and (b) category-wise cross-dataset recall for DoTA and DADA datasets. For each test dataset, the recall values are reported at the probability threshold that yields an overall recall of 80\%.}
    \label{fig:cross_dataset_comparison}
\end{figure*}

As shown in Figure~\ref{fig:cross_dataset_comparison}(a), the DoTA dataset has very few accidents involving cyclists, road obstacles, pedestrians, and motorbikes compared to DADA. We hypothesized that when trained on DoTA and tested on DADA, these categories would have worse recall than others. However, Figure~\ref{fig:cross_dataset_comparison}(b) shows that although the recall is low for cyclists and motorbikes, the recall for pedestrians and obstacles is comparatively high despite their low presence in the training data. Therefore, the under-representation of some object categories does not fully explain the drop in cross-dataset performance.

%Furthermore, differneces in the way the accidents are annotated in the two datasets can contrivute too.

\begin{table}[htbp]
\scriptsize
\caption{Cross-dataset generalization performance on DoTA and DADA. For a given dataset, baseline AP (bAP)$=\frac{\#accident\ videos}{\#all\ videos}$~\cite{saito2015precision}}
\centering
\begin{tabular}{|c|c|c|c|c|}
\hline
\textbf{Train} & \textbf{Test} & \textbf{Method} & \textbf{AP (\%)} & \textbf{mTTA (s)} \\
\hline
\multirow{4}{*}{\centering DoTA ~\cite{yao2022dota}} & \multirow{4}{*}{\shortstack{DADA\\(bAP = 40.8)}} 
& UString (Geometrics) ~\cite{mahmood2023new} & 46.39 & 1.71 \\
& & UString ~\cite{bao2020uncertainty}          & 48.26 & 1.95 \\
& & Graph(Graph)~\cite{thakur2024graph}         & 61.34 & 3.12 \\
& & AAT-DA~\cite{kumamoto2025aat}         & 50.27 & 3.24
\\
& & STAGNet (Ours)                              & \textbf{67.47} & \textbf{3.28} \\
\hline
\multirow{4}{*}{\centering DADA ~\cite{fang2019dada}} & \multirow{4}{*}{\shortstack{DoTA\\(bAP = 50.0)}} 
& UString (Geometrics) ~\cite{mahmood2023new} & 54.85 & 2.45 \\
& & UString ~\cite{bao2020uncertainty}          & 55.63 & 2.17 \\
& & Graph(Graph)~\cite{thakur2024graph}         & 68.55 & 2.86 \\
& & AAT-DA~\cite{kumamoto2025aat}         & 66.23 & 2.18
\\
& & STAGNet (Ours)                              & \textbf{80.15} & \textbf{3.13} \\
\hline
\end{tabular}
\label{tab:cross_dataset_generalization}
\end{table}

\subsubsection{Impact of Global Features}

We analyze the impact of various global feature extraction methods on our accident anticipation framework. Since spatial features extracted from pretrained image-based networks such as VGG16 focus solely on individual frames, they lack the ability to capture the temporal dynamics across frames. To overcome this limitation, spatio-temporal features extracted using I3D and SlowFast networks were explored. As shown in Table~\ref{tab:global_feature_comparison}, the I3D model improves performance over VGG16. Furthermore, our proposed use of SlowFast yields the best performance, outperforming both VGG16 and I3D. This improvement is attributed to the dual-pathway design of SlowFast, which effectively captures both fast motion cues and slow-evolving spatial context, providing a more comprehensive spatio-temporal representation. %These capabilities make SlowFast highly effective at predicting future accidents.

Note that we have kept the object detector (YOLO), object-level feature extractor (VGG16), and the global feature extractor (SlowFast) frozen at their pre-trained weights during training. All prior works known to us have also frozen their object detectors and feature extractors. This is reasonable for the YOLO and VGG16 models, as they have been trained on large-scale datasets containing vehicles, pedestrians, and other road objects. However, the SlowFast networks have been trained on human activity datasets~\cite{Feichtenhofer2018SlowFastNF}. Therefore, finetuning them on road traffic data may improve the models performance in accident anticipation. To evaluate this, we finetuned the SlowFast networks on the corresponding training sets of DoTA and DADA. However, there was only a marginal gain in AP and no improvement in mTTA, as evident from the results in Table~\ref{tab:global_feature_comparison}.

\begin{table}[htbp]
\caption{Comparison of the effects of different global feature extractors on our STAGNet model.}
\centering
\scriptsize
\begin{tabular}{|c|c|c|c|}
\hline
\textbf{Dataset} & \textbf{Method} & \textbf{AP (\%)} & \textbf{mTTA (s)} \\
\hline
\multirow{3}{*}{DoTA ~\cite{yao2022dota}} 
                      & VGG16 ~\cite{simonyan2014very}    & 80.89 & 3.06 \\
                      & I3D   ~\cite{carreira2017quo}    & 88.52 & 2.98 \\
                      & SlowFast ~\cite{Feichtenhofer2018SlowFastNF}  & \textbf{92.73} & \textbf{3.23} \\
                      & SlowFast (finetuned)  & \textbf{93.24} & 3.19 \\
\hline
\multirow{3}{*}{DADA ~\cite{fang2019dada}} 
                      & VGG16 ~\cite{simonyan2014very}     & 81.94 & 2.96 \\
                      & I3D ~\cite{carreira2017quo}       & 89.40 & 2.92 \\
                      & SlowFast ~\cite{Feichtenhofer2018SlowFastNF}  & \textbf{95.39} & \textbf{3.08} \\
                      & SlowFast (finetuned)  & \textbf{95.81} & 3.01 \\
\hline
\end{tabular}
\label{tab:global_feature_comparison}
\end{table}

\begin{table}[h]
\centering
\scriptsize
\caption{Performance of UString with global features extracted from SlowFast networks.}
\begin{tabular}{|c|c|c|c|c|}
\hline
\textbf{Dataset} & \textbf{Method} & \textbf{Global Feature} & \textbf{AP (\%)} & \textbf{mTTA (s)} \\
\hline
\multirow{2}{*}{DoTA~\cite{yao2022dota}} 
& \multirow{2}{*}{UString~\cite{bao2020uncertainty}} 
& VGG16~\cite{simonyan2014very} & 75.40 & 2.26 \\
& & SlowFast~\cite{Feichtenhofer2018SlowFastNF} & 86.38 & 1.90 \\
\hline
\multirow{2}{*}{DADA~\cite{fang2019dada}} 
& \multirow{2}{*}{UString~\cite{bao2020uncertainty}} 
& VGG16~\cite{simonyan2014very} & 63.95 & 1.03 \\
& & SlowFast~\cite{Feichtenhofer2018SlowFastNF} & 66.14 & 1.11 \\
\hline
\end{tabular}
\label{tab:ustring_slowfast}
\end{table}

\noindent
We also evaluated the UString framework by integrating global features extracted from the SlowFast networks. The results, as presented in Table~\ref{tab:ustring_slowfast}, show a notable improvement in performance compared to the original UString method, which used VGG16 global features. This demonstrates that incorporating global context through SlowFast features enhances the UString model as well. %Specifically, the AP increased to 86.38 on DoTA and to 66.14 on DADA, while achieving competitive mTTA values of 1.90s and 1.11s, respectively. 

\subsubsection{Ablation Study}

\begin{table}[htbp]
\scriptsize
\setlength{\tabcolsep}{4pt}
\caption{Results of the ablation study on the DADA and DoTA datasets. Each model variant removes one or more components (LSTM, GATv2, or SlowFast) to analyze their individual impact. Note: When GATv2 was not used, GCN layers were used. When SlowFast features were not used, I3D features were used.}
\centering
\begin{tabular}{|c|c|c|c|c|c|}
\hline
\textbf{Data} & \textbf{LSTM} & \textbf{GATv2} & \textbf{SlowFast} & \textbf{AP (\%)} & \textbf{mTTA (s)} \\
\hline
\multirow{7}{*}{DoTA ~\cite{yao2022dota}} 
 & \ding{55} & \ding{55} & \ding{55} & 83.50 & 3.14 \\
 & \ding{51} & \ding{55} & \ding{55} & 89.98 & 3.06 \\
 & \ding{55} & \ding{51} & \ding{55} & 85.89 & 3.13 \\
 & \ding{55} & \ding{55} & \ding{51} & 90.03 & 2.97 \\
 & \ding{51} & \ding{51} & \ding{55} & 88.52 & 2.98 \\
 & \ding{51} & \ding{55} & \ding{51} & 92.18 & 3.12 \\
 & \ding{55} & \ding{51} & \ding{51} & 90.75 & 3.13 \\
 & \ding{51} & \ding{51} & \ding{51} & \textbf{92.73} & \textbf{3.23} \\
\hline
\multirow{7}{*}{DADA ~\cite{fang2019dada}} 
 & \ding{55} & \ding{55} & \ding{55} & 84.04 & 2.92 \\
 & \ding{51} & \ding{55} & \ding{55} & 89.18 & 2.89 \\
 & \ding{55} & \ding{51} & \ding{55} & 84.65 & 2.96 \\
 & \ding{55} & \ding{55} & \ding{51} & 92.95 & 2.89 \\
 & \ding{51} & \ding{51} & \ding{55} & 89.40 & 2.92 \\
 & \ding{51} & \ding{55} & \ding{51} & 94.75 & 3.04 \\
 & \ding{55} & \ding{51} & \ding{51} & 93.51 & 2.79 \\
 & \ding{51} & \ding{51} & \ding{51} & \textbf{95.39} & \textbf{3.08} \\
\hline
\end{tabular}
\label{tab:ablation_combined}
\end{table}

The ablation results presented in Table~\ref{tab:ablation_combined} demonstrate the individual contributions of each additional component or modification — LSTM, GATv2, and SlowFast — to the overall performance of our accident anticipation framework.

From the results, it is evident that reverting any of the modifications causes a drop in average precision (AP) on both datasets. However, using SlowFast instead of I3D features and/or adding an LSTM network in global feature learning has a much larger impact than the addition of GATv2 in object graph learning. For instance, on the DoTA dataset, the baseline without any of the three modifications achieves an AP of 83.50\%. Adding the LSTM alone increases AP considerably to 89.98\%, indicating the critical role of temporal modeling. Similarly, replacing I3D with SlowFast as the feature extractor itself yields an AP of 90.03\%, highlighting the advantage of powerful spatio-temporal feature representations. Combining LSTM and SlowFast further improves performance to 92.18\%, and the best results are achieved when all three components (LSTM, GATv2, and SlowFast) are used together, reaching 92.73\% AP and an average mTTA of 3.23 seconds. A similar trend is observed with the DADA dataset. The baseline without any modifications yields 84.04\% AP, which is improved to 89.18\% by adding the LSTM alone. Replacing I3D with SlowFast features as the only modification results in 92.95\% AP. Again, the highest performance, with 95.39\% AP and 3.08 s mTTA, is attained when all components are used. 

The ablation study also reveals that reverting either modification reduces mTTA. Furthermore, we note that neither modification, when applied in isolation, improves the mTTA compared to the baseline Graph(Graph) model.

Overall, the ablation study demonstrates that using SlowFast features and/or adding an LSTM module contributes to anticipating accidents much more precisely, while using GATv2 in place of GCN layers in the object graph learning module provides marginal gains. In terms of mTTA, all three modifications are required to anticipate accidents earlier.

\subsubsection{Analysis of Computational Efficiency and Execution Speed}

Tables~\ref{tab:execution_fps} and~\ref{tab:execution_flops} show the execution times and the number of floating-point operations (FLOPs), respectively, for the different stages of the pipeline across different methods. These experiments were conducted on a computing platform with the following configuration: GPU – Tesla P100 with 16 GiB of memory, CPU – 2-core Intel Xeon (at 2.2 GHz), and 29 GiB of RAM. 

According to the results, object detection and VGG16 feature extraction from bounding boxes (19 most confident detections) account for the majority of the computational cost: 0.063 s/frame, 322.46 GFLOPs/frame. This is a burden common to all compared methods, except for our pruned variant, STAGNet-Lite. In terms of global feature extraction, VGG is the fastest, followed by I3D and SlowFast, respectively. In terms of model inference time (excluding pre-trained feature extractors), STAGNet and Graph(Graph) operate faster than the BNN-based Ustring model and the transformer-based AAT-DA model. Overall, UString and Graph(Graph) models using either VGG16 or I3D global features operate at close to 10 FPS, while our STAGNet model with SlowFast features is slightly slower at 9 FPS. However, we note that the implementations of neither STAGNet nor other models have been optimized for efficiency. In fact, none of the comparison methods used in this work have previously reported run-times. Therefore, we expect that these models could be further optimized. Especially, the time-consuming object feature extraction can be parallelized to improve speed.

On the other hand, our pruned model, STAGNet-Lite, has achieved 21 FPS, which is much faster than all other models. It also maintains high AP and mTTA values, falling slightly short of the full STAGNet model but higher than the remaining models for ego-involved accidents (DoTA and DADA datasets). However, it shows a more prominent drop in AP compared to the full model for non-ego-involved accidents (DAD dataset), which is expected due to pruning the modules responsible for object detection and object-object interaction modeling. Nevertheless, STAGNet-Lite can be practically useful for resource-constrained ADAS environments due to its execution speed and good performance in anticipating ego-involved accidents, which is the most important class of accidents.

\begin{table}[htbp]
\scriptsize
\caption{Execution times and speed of different methods. Times for different stages are reported in seconds per frame (s/f), averaged over DoTA and DADA datasets. The overall inference speed is reported as frames per second (FPS). Note that the inference time of AAT-DA shows the time taken to generate its driver attention maps as well.}
\centering
\setlength{\tabcolsep}{5pt}
\renewcommand{\arraystretch}{1.2}
\begin{tabular}{|c|c|c|c|c|}
\hline
\shortstack{\textbf{Method}} &
\shortstack{\textbf{Object}\\\textbf{ Detection}\\\textbf{\& Feature}\\\textbf{Extraction (s/f)}} &
\shortstack{\textbf{Global Feature}\\\textbf{Extraction (s/f)}} &
\shortstack{\textbf{Inference}\\\textbf{(s/f)}} &
\shortstack{\textbf{Overall}\\\textbf{speed}\\\\\textbf{(FPS)}} \\
\hline
UString            & \multirow{4}{*}{0.063} & 0.025 (VGG16)    & 0.012 & 9.960 \\
Graph(Graph)       &                        & 0.032 (I3D)      & 0.002 & 10.199 \\
AAT-DA       &                              & 0.025 (VGG16)    & 0.009+0.085* & 5.464 \\
STAGNet      &                              & 0.032 (I3D)      & 0.002 & 10.197 \\
STAGNet  &                                  & 0.045 (SlowFast) & 0.002 & 9.059 \\
\hline
STAGNet-Lite  &    -                   & 0.045 (SlowFast) & 0.002 & 21.317 \\
\hline
\end{tabular}
\label{tab:execution_fps}
\begin{flushleft}
\scriptsize \textit{*} Time for saliency/driver-attention computation. 
\end{flushleft}
\vspace{-4pt}
\end{table}

\begin{table}[htbp]
\scriptsize
\caption{FLOP counts of different methods. Counts for different stages are reported in GFLOPs/frame. The counts were obtained using the Python library THOP~\cite{THOP}.}
\centering
\setlength{\tabcolsep}{5pt}
\renewcommand{\arraystretch}{1.2}
\begin{tabular}{|c|c|c|c|c|}
\hline
\multirow{2}{*}{\textbf{Method}} & \multicolumn{4}{c|}{\textbf{GFLOPs}} \\ \cline{2-5}
 & \shortstack{\textbf{Object Detection}\\\textbf{\& Feature}\\ \textbf{Extraction}} 
 & \shortstack{\textbf{Global Feature}\\\textbf{Extraction}} 
 & \shortstack{\textbf{Model}\\\textbf{Inference}} 
 & \textbf{Total} \\
\hline

UString & \multirow{4}{*}{322.46} & 15.47 (VGG16)    
& 10.49   & 348.42 \\
Graph (Graph) & & 41.75 (I3D)           
& 2.14   & 366.35 \\
AAT-DA        &                        & 15.47 (VGG16)       
& 1.05  & 461.96* \\
STAGNet       &                        & 50.58 (SlowFast)     
& 2.20   & 375.24 \\
\hline
STAGNet-Lite      &       -         & 50.58 (SlowFast)     
& 0.17   & 50.75 \\
\hline
\end{tabular}
\label{tab:execution_flops}
\begin{flushleft}
\scriptsize \textit{*} The total includes the 122.98 GFLOPs used for saliency/driver-attention computation. 
\end{flushleft}
\vspace{-4pt}
\end{table}

\section{Conclusion}

In this work, we demonstrate that state-of-the-art graph neural networks for accident anticipation can be substantially improved by using SlowFast networks to extract global, spatio-temporal features of video frames and aggregating them through an LSTM network. Our experiments on benchmark datasets show that the proposed STAGNet model can anticipate ego-involved accidents earlier and more accurately compared to existing methods. Our approach outperforms the others even when tested on a dataset different from the one used for training (cross-dataset evaluation). However, for all algorithms tested, the cross-dataset average precision and mean time-to-accident are notably lower than those achieved when cross-validated on the same dataset. The reasons for this drop in performance are not clear from the experiments we conducted. This suggests that further research is needed to investigate and improve the generalization of these models to new environments and traffic conditions.  

We note that our full STAGNet model and the other models compared here are currently not optimized for execution speed. Our pruned model, STAGNet-Lite, operates at a much faster speed while being almost as effective as the full model in detecting ego-involved accidents. In future work, we will further explore techniques to make the models more computationally efficient. We also plan to enhance vehicle localization by moving beyond bounding box–based approaches, which can leave empty spaces and fail to encapsulate vehicles tightly. Utilizing segmentation techniques would enable more precise delineation of vehicle boundaries, thereby improving detection accuracy. However, this could lead to a reduction in inference speed. Therefore, the trade-offs need to be considered carefully. Additionally, incorporating depth information or utilizing the cameras' intrinsic matrix could enable more accurate estimation of the distance between vehicles, facilitating better scene understanding and potentially improving the robustness of accident prediction models.

%%%%%%%%%%%%%%   Bibliography   %%%%%%%%%%%%%%
\normalsize
\bibliography{VisionCW}

%%%%%%%%%%%%  Supplementary Figures  %%%%%%%%%%%%
%\clearpage

%%%%%%%%%%%%%%%%   End   %%%%%%%%%%%%%%%%
%\end{multicols}  % Method B for two-column formatting (doesn't play well with line numbers), comment out if using method A
\end{document}